%% file: main.tex
\documentclass[11pt,a4paper]{article}
\usepackage[utf8]{inputenc}

\usepackage[hyperref]{emnlp-ijcnlp-2019}
\usepackage{times}
\usepackage{latexsym}
\usepackage{booktabs}
\usepackage{multirow}
\usepackage{adjustbox}
\usepackage{amssymb}
\usepackage{url}
\usepackage{amsmath}
\usepackage{enumitem}

\aclfinalcopy % Uncomment this line for the final submission

\newcommand{\embedim}{\Phi}
\newcommand{\embedtext}{E}
\newcommand{\embedpassage}{\Psi}
\newcommand{\embedpos}{\pi}

\title{Fusion of Detected Objects in Text for Visual Question Answering}

\author{Chris Alberti
\;\;\;\;
  Jeffrey Ling\thanks{\hspace{0.1em} Work done as part of the Google AI residency.}
\;\;\;\;
  Michael Collins 
\;\;\;\;
  David Reitter \\
Google Research\\
{ \small   \texttt{\{chrisalberti, jeffreyling, mjcollins, reitter\}@google.com}}}

\date{}

\begin{document}
\maketitle
\begin{abstract}

To advance models of multimodal context, we introduce a simple yet powerful neural architecture for data that combines vision and natural language. The ``Bounding Boxes in Text Transformer'' (B2T2) also leverages referential information binding words to portions of the image in a single unified architecture.  B2T2 is highly effective on the Visual Commonsense Reasoning benchmark\footnote{\url{https://visualcommonsense.com}}, achieving a new state-of-the-art with a 25\% relative reduction in error rate compared to published baselines and obtaining the best performance to date on the public leaderboard  (as of May 22, 2019). A detailed ablation analysis shows that the early integration of the visual features into the text analysis is key to the effectiveness of the new architecture. A reference implementation of our models is provided\footnote{\url{https://github.com/google-research/language/tree/master/language/question_answering/b2t2}}.

\end{abstract}

\section{Introduction}

It has long been understood that the meaning of a word is systematically and predictably linked to the context in which it occurs (e.g., \citealt{firth1957synopsis,harris1954distributional,deerwester1990indexing,mikolov2013distributed}).  Different notions of context have resulted in different levels of success with downstream NLP tasks. Recent neural architectures including Transformer \citep{vaswani2017attention} and BERT \citep{devlin2018bert} have dramatically increased our ability to include a broad window of potential lexical hints.  However, the same capacity allows for multimodal context, which may help model the meaning of words in general, and also sharpen its understanding of instances of words in context (e.g., \citealt{bruni2014multimodal}).  

\begin{figure}
  \includegraphics[width=\columnwidth]{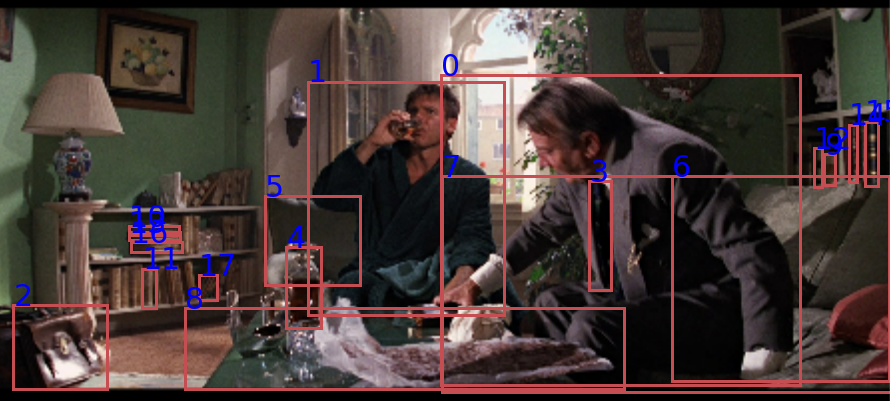}
\begin{scriptsize}
{\setlength{\tabcolsep}{2pt}
\begin{tabular}{cp{6.5cm}} \\
Q: & What was [1] doing before he sat
   in his living room? \vspace{1mm} \\
A$_1$: & He was reading [10]. \\
A$_2$: & He was taking a shower. \checkmark \\
A$_3$: & [0] was sleeping until the noise 
    [1] was making woke him up.\\
A$_4$: & He was sleeping in his bedroom.
   \vspace{1mm} \\
R$_1$: & His clothes are disheveled and his
    face is glistening like he's sweaty. \\
R$_2$: & [0] does not look wet yet, but [0]
    looks like his hair is wet, and bathrobes
    are what you wear before or after a shower. \\
R$_3$: & He is still wearing his bathrobe. \checkmark\\
R$_4$: & His hair appears wet and there is clothing
    hanging in front of him on a line as if to dry.
\end{tabular}}
\end{scriptsize}
\caption{An example from the VCR dataset. The tasks consists in picking an answer A$_{1-4}$, and then picking a rationale R$_{1-4}$. The data contains explicit pointers in the text to bounding boxes in the image.}
\label{fig:rationale-needed}
\end{figure}

In this paper, we consider visual context in addition to language and show that the right integration of visual and linguistic information can yield improvements in visual question answering. The challenge we consider is to answer natural-questions related to a given image. The more general question we address in the context of this problem is how to encode visual and verbal information in a neural architecture. How to best do that is still unclear.  How are text entities bound to objects seen in images?  Are text and image best integrated late, allowing for independent analysis (\emph{late fusion}), or should the processing of one be conditioned on the analysis of the other (\emph{early fusion})?  How is cross-modal co-reference best encoded at all?  Does it make sense to ground words in the visual world before encoding sentence semantics?
% \cite{snoek2005early}

In this work we gather evidence to answer these questions by designing the Bounding  Boxes  in  Text  Transformer, B2T2 for short, a neural architecture for multimodal encoding of natural language and images, and we evaluate B2T2 on the Visual Commonsense Reasoning benchmark (VCR, \citealt{zellers2019}).

Figure \ref{fig:rationale-needed} shows an illustrative example from the VCR benchmark.
VCR is well suited to test rich multimodal representations because it requires the analysis of images depicting people engaged in complex activities; it presents questions, answers and rationales created by human annotators rather than automatic generation; it has a clean multiple-choice interface for evaluation; and yet it is still challenging thanks to a careful selection of answer choices through adversarial matching. VCR has much longer questions and answers compared to other popular Visual Question Answering (VQA) datasets, such as VQA v1 \citep{VQA}, VQA v2 \citep{balanced_vqa_v2} and GQA \citep{hudson2019gqa}, requiring more modeling capacity for language understanding.

In our experiments, we found that early fusion of co-references between textual tokens and visual features of objects was the most critical factor in obtaining improvements on VCR. We found that the more visual object features we included in the model's input, the better the model performed, even if they were not explicitly co-referent to the text, and that positional features of objects in the image were also helpful. We finally discovered that our models for VCR could be trained much more reliably when they were initialized from pretraining on Conceptual Captions \citep{sharma2018conceptual}, a public dataset of about 3M images with captions. From the combination of these modeling improvements, we obtained a new model for visual question answering that achieves state-of-the-art on VCR, reducing  error rates by more than 25\% relative to the best published and documented model \citep{zellers2019}.

\section{Problem Formulation}

\begin{table}
\begin{scriptsize}
{\setlength{\tabcolsep}{2pt}
\begin{tabular}{ccp{4cm}} \\
Symbol & Type & Description \\ \midrule

$m$ & $\mathbb{N}$ & number of extracted bounding boxes \\

$n$ & $\mathbb{N}$ & number of tokens input to BERT \\

$k$ & $\mathbb{N}$ & number of positional embeddings for image coordinates, usually 56 \\

$d$ & $\mathbb{N}$ & visual features dimension, usually 2048 \\

$h$ & $\mathbb{N}$ & hidden dimension of BERT, usually 1024 \\

$l$ & $\{0, 1\}$ & a binary label \\

$I$ & $\mathbb{R}^{\cdot \times \cdot \times 3}$ & an image \\

$B$ & $\mathbb{R}^{m \times 4}$ & rectangular bounding boxes on $I$, as coordinates of opposite corners \\

$R$ & $\{0, 1\}^{m \times n}$ & matrix encoding which bounding boxes in $B$ correspond to which tokens in $T$ \\

$T$ & $\mathbb{N}^{n \times 2}$ & input tokens, each expressed as word piece id and token type \\

$\embedim$ & $\mathbb{R}^{\cdot \times \cdot \times 3} \rightarrow \mathbb{R}^d$ & a function to extract visual feature vectors from an image \\

$\embedpos$ & $\mathbb{R}^4 \rightarrow \mathbb{R}^d$ & a function to embed the position and shape of a bounding box \\

$\embedpassage$ & 
$\mathbb{R}^{n \times h} \rightarrow \mathbb{R}^{h}$ & a function to compute a passage embedding from per-token embeddings \\

$\embedtext$ & $\mathbb{N}^{n \times 2} \rightarrow \mathbb{R}^{n \times h}$ & non-contextualized token embeddings, encoding word piece ids, token types and positions \\ \bottomrule
\end{tabular}}
\end{scriptsize}
\caption{Glossary of mathematical symbols used in this work.}
\label{tab:glossary}
\end{table}

% TODO: MAKE SURE THERE IS NO FALSE OVERLAP WITH THE VCR SPEC BELOW
In this work, we assume data comprised of 4-tuples $(I, B, T, l)$ where
\begin{enumerate}
    \item $I$ is an image,
    \item $B = [b_1, \ldots, b_m]$ is a list of bounding boxes referring to regions of $I$, where each $b_i$ is identified by the lower left corner, height and width,
    \item $T = [t_1, \ldots, t_n]$ is a passage of tokenized text, with the peculiarity that some of the tokens are not natural language, but explicit references to elements of $B$, and
    \item $l$ is a binary label in $\{0, 1\}$.
\end{enumerate}

While it might seem surprising to mix natural text with explicit references to bounding boxes, this is actually a quite natural way for people to discuss objects in images and the VCR dataset is annotated in exactly this way.

We assume an image representation function $\embedim$ that converts an image, perhaps after resizing and padding, to a fixed size vector representation of dimension $d$.

We similarly assume a pretrained textual representation capable of converting any tokenized passage of text, perhaps after truncating or padding, into a vector representation of dimension $h$. We assume a context independent token representation $\embedtext$ in the shape of a vector of dimension $h$ for each token and a passage level representation $\embedpassage$ which operates on $\embedtext(T)$ and returns a passage level vector representation of dimension $h$.

We refer the reader to Table \ref{tab:glossary} for an overview of the notation used in this work. Full details on how the VCR dataset is encoded into this formalism are given in Section \ref{sec:data}.

\section{Models and Methods}

\begin{figure*}[t]% keep both figures on same page
\begin{minipage}{1.0\columnwidth}
\includegraphics[width=\columnwidth]{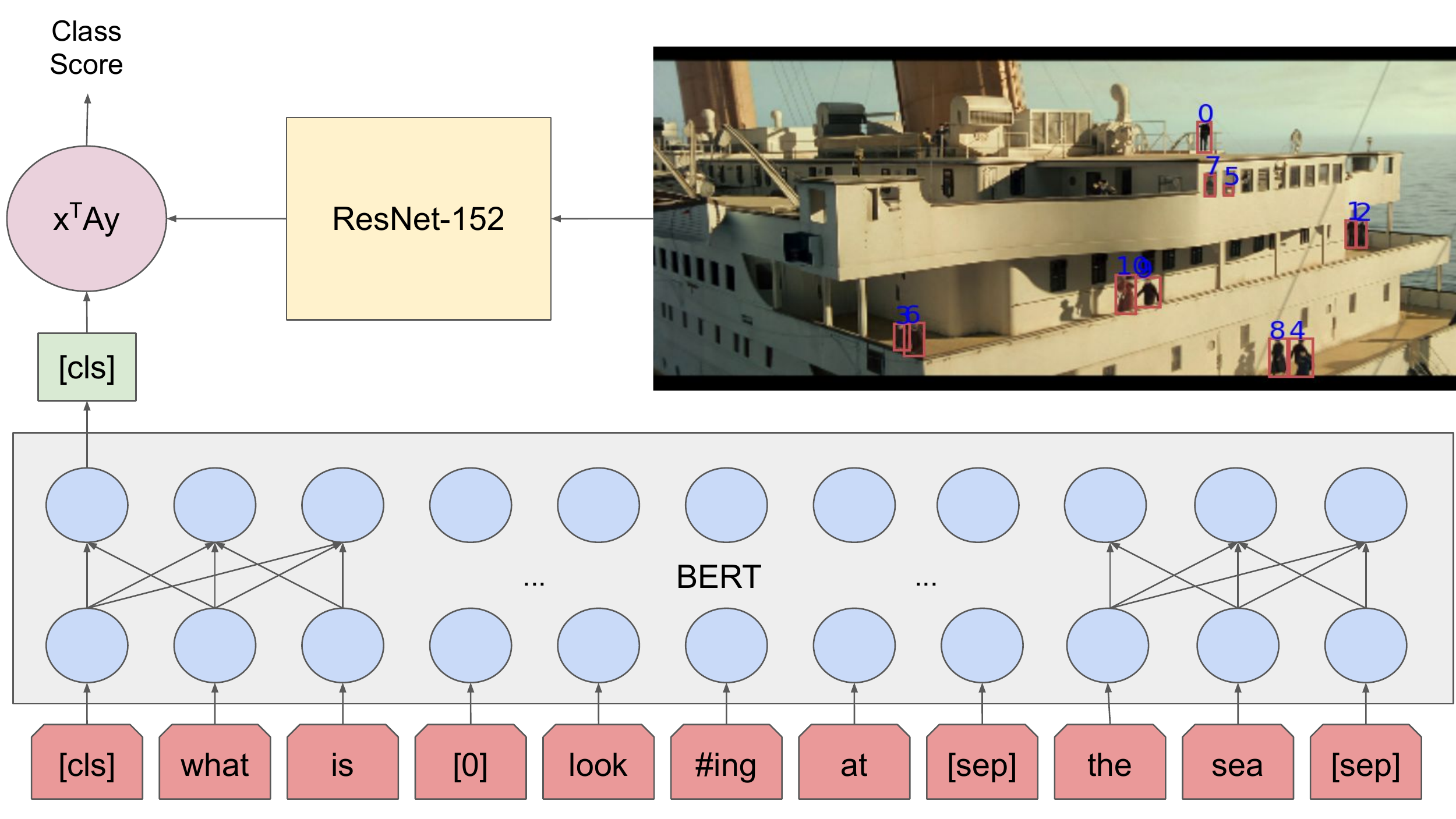}
\caption{Dual Encoder architecture with late fusion. The model extracts a single visual feature vector from the entire image. Bounding boxes are ignored.
}
\label{fig:dual-encoder}
\end{minipage}\hspace{3\tabcolsep}%
\begin{minipage}{1.0\columnwidth}
\includegraphics[width=\columnwidth]{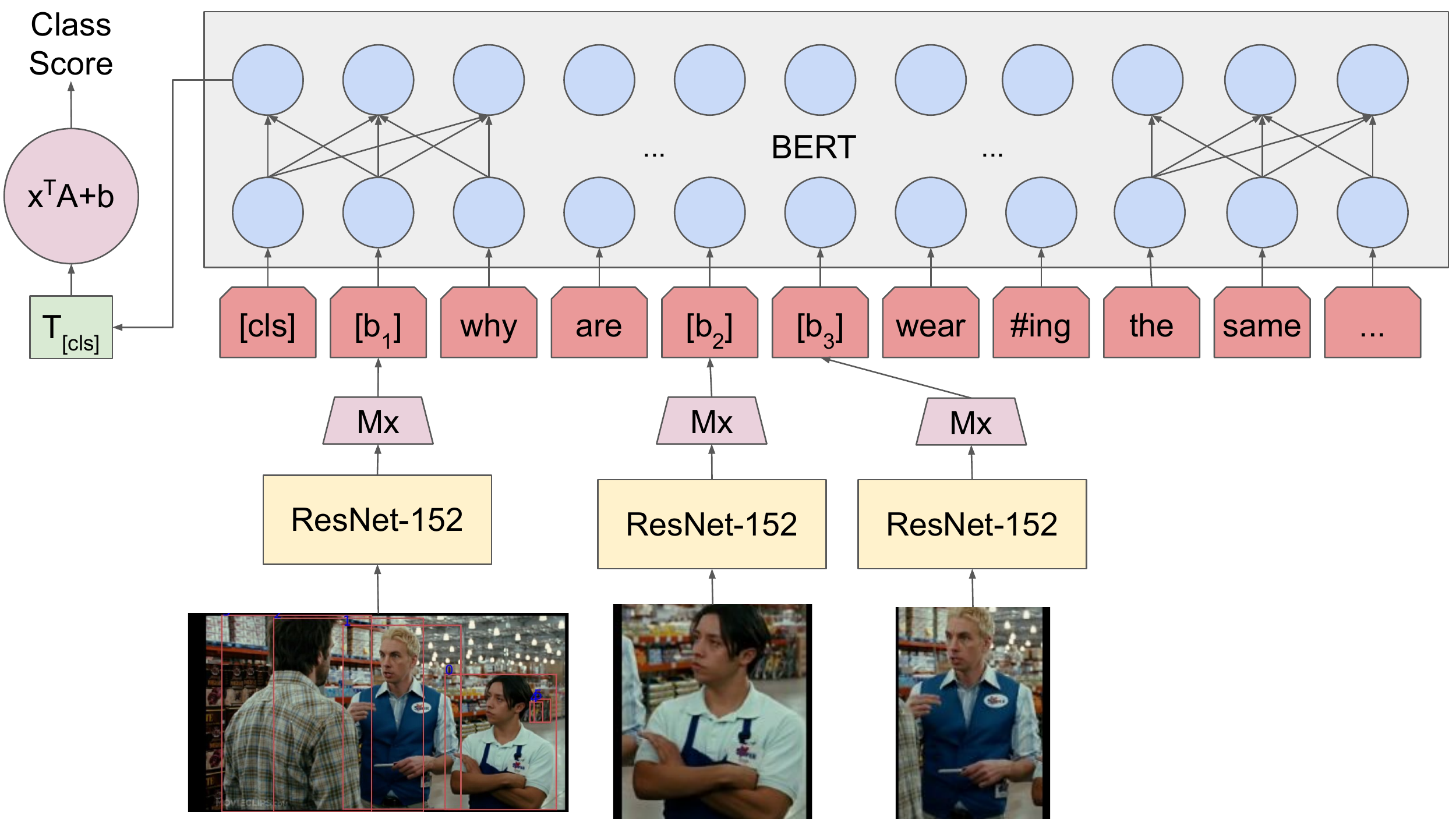}
\caption{B2T2 architecture with early fusion. Bounding boxes are inserted where they are mentioned in the text and at the end of the input, as described in Sec.\ \ref{sec:data}.
}
\label{fig:deep-bottom-up}
\end{minipage}
\end{figure*}

We evaluate two main architectures: ``Dual Encoder'', a late fusion architecture where image and text are encoded separately and answer scores are computed as an inner product, and the full B2T2 model, an early fusion architecture where visual features are embedded on the same level as input word tokens.  Section~\ref{sec:ablations} will summarize experiments with model variants to answer the research questions laid out in the introduction and to analyze what works, and why.

\subsection{Dual Encoder}

Dual Encoders, discussed for example by \citet{wu2018starspace} and \citet{gillick2018end}, are models that embed objects of potentially different types into a common representation space where a similarity function can be expressed e.g.\ as a dot product or a cosine similarity. A notable example of a dual encoder for image classification is WSABIE, proposed by \citet{weston2011wsabie}.

Our Dual Encoder architecture is shown in Figure \ref{fig:dual-encoder}. We model the class distribution as
\[
   p(l = 1 | I, T) = \frac 1 {1 + e^{-\embedpassage(\embedtext(T))^\top D \embedim(I)}}
\]
where $D$ is a learned matrix of size $d \times h$. In this model, co-reference information is completely ignored, and the model must rely on fixed dimensional vectors for the late fusion of textual and visual contexts. However, we found this to be surprisingly competitive on VCR compared to published baselines, perhaps due to our choice of powerful pretrained models.

%{\color{red} DR: Is there anything in the literature that suggests this dual encoder architecture would fare well on this task?  If we had reasons beyond intuition to try this out, we should mentioned them here. I'll add to the intro of section 3 with that in mind. CA: Maybe retrieval models -- perhaps Kenton's ORQA model works like this. }

\subsection{B2T2}

Our B2T2 architecture is shown in Figure \ref{fig:deep-bottom-up}. 
We model the class distribution as
\[
  p(l|I, B, R, T) = \frac {e^{\embedpassage(\embedtext'(I, B, R, T)) \cdot a_l + b_l}}
         {\sum_{l'} e^{\embedpassage(\embedtext'(I, B, R, T)) \cdot a_{l'} + b_{l'}}}
\]
where $a_l \in \mathbb{R}^h$ and $b_l \in \mathbb{R}$ for $l \in \{0, 1\}$ are learned parameters. $\embedtext'(I, B, R, T)$ is a non-contextualized representation for each token and of its position in text, but also of the content and position of the bounding boxes. The key difference from ``Dual Encoder'' is that text, image and bounding boxes are combined at the level of the non-contextualized token representations rather than right before the classification decision.

\begin{figure}[t]
\includegraphics[width=\columnwidth,trim={0 2cm 0 2cm},clip]{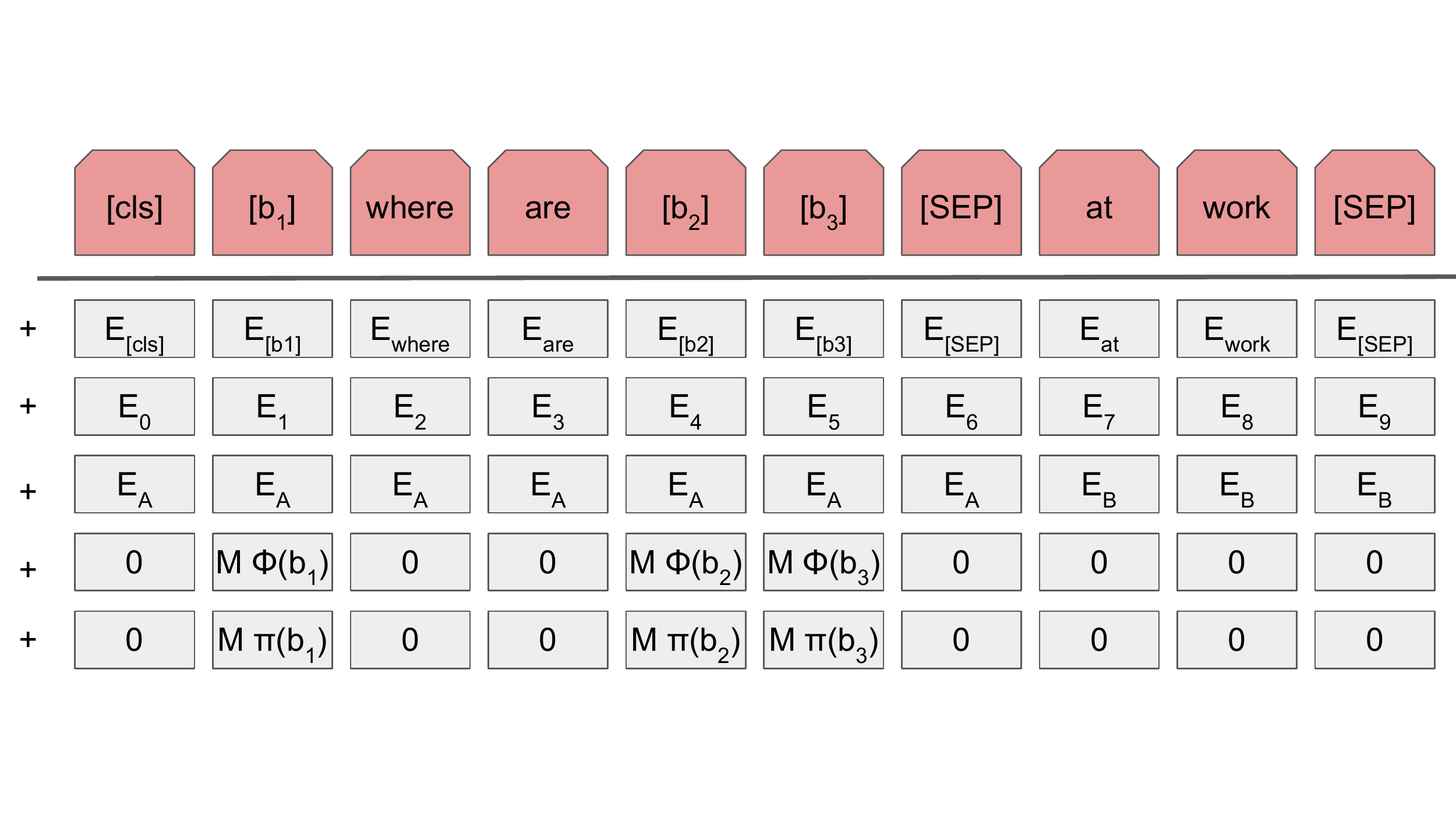}
\caption{How input embeddings are computed in our B2T2 architecture.}
\label{fig:embedding}
\end{figure}

The computation of $\embedtext'(I, B, R, T)$ is depicted in Figure \ref{fig:embedding}. More formally, for a given example, let matrix $R \in \{0, 1\}^{m \times n}$ encode the references between the bounding boxes in $B$ and the tokens in $T$, so that $R_{ij}$ is $1$ if and only if bounding box $i$ is referenced by token $j$. Then
\begin{align*}
& \embedtext'(I, B, R, T) = \\
& \;\; \embedtext(T) + \sum_{i=1}^m 
  R_i
   \left[ M (\embedim(\mbox{crop}(I, b_i)) + \embedpos(b_i))
   \right]^\top
\end{align*}
where $M$ is a learned $h \times d$ matrix, $\embedim(\mbox{crop}(I, b_i))$ denotes cropping image $I$ to bounding box $b_i$ and then extracting a visual feature vector of size $d$, and $\embedpos(b_i)$ denotes the embedding of $b_i$'s shape and position information in a vector of size $d$.

To embed the position and size of a bounding box $b$, we introduce two new learnable embedding matrices $X$ and $Y$ of dimension $k \times \frac d 4$. Let the coordinates of the opposite corners of $b$ be $(x_1, y_1)$ and $(x_2, y_2)$, after normalizing so that a bounding box covering the entire image would have $x_1 = y_1 = 0$ and $x_2 = y_2 = k$. Position embeddings are thus defined to be
\[
  \pi(b) = \mbox{concat}(X_{\lfloor x_1 \rfloor}, Y_{\lfloor y_1 \rfloor},
                         X_{\lfloor x_2 \rfloor}, Y_{\lfloor y_2 \rfloor})
\]

\subsection{Loss}
\label{sec:loss}

All of our models are trained with binary cross entropy loss using label $l$. Denoting $p := p(l=1|I,B,R,T)$, we have for each example
\begin{equation*}
\mathcal{L_{\text{BCE}}} = l \log p + (1 - l) \log (1 - p)
\end{equation*}

\subsection{Pretraining on Conceptual Captions}

\begin{figure}[t]
\includegraphics[width=\columnwidth]{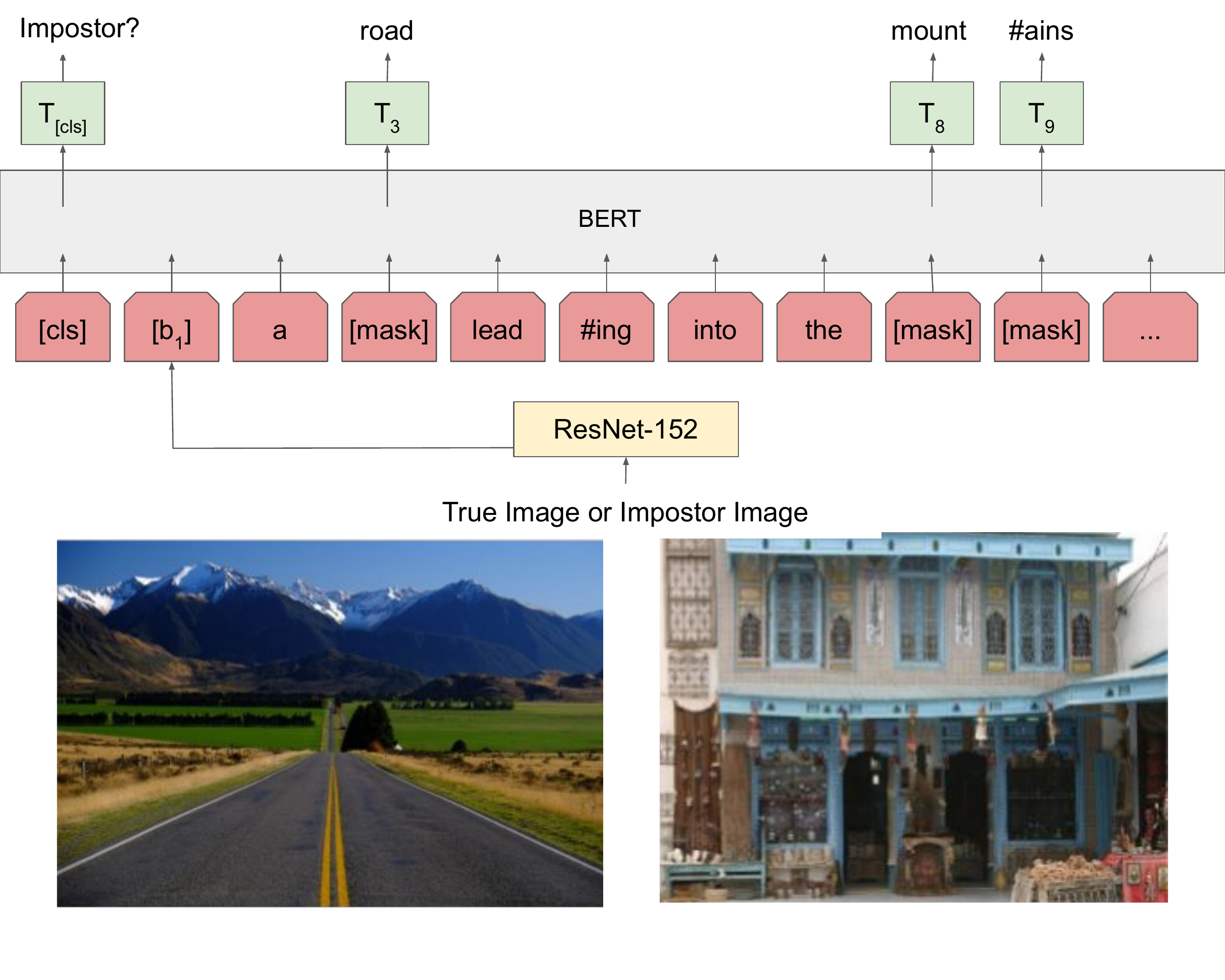}
\caption{Mask-LM pretraining for B2T2.}
\label{fig:mask-lm-pretraining}
\end{figure}

Before training on VCR, we pretrain B2T2 on image and caption pairs using a Mask-LM pretraining technique like the one used in BERT \citep{devlin2018bert}. The setup used during pretraining is shown in Figure \ref{fig:mask-lm-pretraining}, where the model uses the image as additional context when filling in the mask.

We use two tasks for pretraining: (1) impostor identification and (2) masked language model prediction.
For the impostor task, we sample a random negative caption for each image and ask the model to predict whether the caption is correctly associated.
For mask-LM, we randomly replace tokens in the caption with the \texttt{[MASK]} token, and the model must predict the original token (see \citet{devlin2018bert} for more details).

Formally, the pretraining data consist of images $I$ and captions $T$. We do not consider bounding boxes during pretraining, so $B = \emptyset$. The binary label $l$ indicates whether the caption is an impostor or not.
The loss for impostor identification is binary cross entropy $\mathcal{L}_{\text{BCE}}$ with label $l$ as in~\ref{sec:loss}. 
We denote the loss for mask-LM as $\mathcal{L}_{\text{MLM}}$, which is the summed cross entropy of the predicted token distributions against the true tokens.

To ensure that our model correctly grounds the language to the image with the mask LM loss, we only use it for positive captions, zeroing it out for negative captions.
Our final objective is the sum of the losses:
\[ \mathcal{L} = \mathcal{L}_{\text{BCE}} + I[l = 1] \cdot \mathcal{L}_{\text{MLM}} \]
where $I[l=1]$ is an indicator for the label $l$ being positive for the image and caption pair.

We pretrain on Conceptual Captions \citep{sharma2018conceptual}, a dataset with over 3M images paired with captions.\footnote{We also tried pretraining on MS-COCO images and captions \citep{lin2014microsoft}, but found this to be ineffective. This could be because MS-COCO is smaller (with around 80k images, 400k captions).
}
We found empirically that pretraining improves our model slightly on VCR, but more importantly, allows our model to train stably. Without pretraining, results on VCR exhibit much higher variance. We refer the reader to Section \ref{sec:ablations} for an ablation analysis on the effect of pretraining.

\subsection{Implementation Details}
\label{sec:implementation}

We use ResNet-152\footnote{Publicly available at \url{tfhub.dev}} \cite{he2016identity} pretrained on ImageNet for $\embedim$, which yields a vector representation of size $d=2048$.  BERT-Large \citep{devlin2018bert} provides both $\embedtext$ and $\embedpassage$. The latter is a pretrained Transformer with 24 layers, 16 attention heads, and hidden size 1024. For BERT, $\embedtext$ corresponds to its token embeddings, $\embedpassage$ to the \texttt{[CLS]} token representation in the final layer, and so $\embedpassage(\embedtext(T))$ corresponds to the BERT passage representation of size $h=1024$.

We found empirically that it was slightly better to keep $\embedim$ fixed rather than fine-tuning it, but that it was of critical importance to fine-tune $\embedpassage$ and $\embedtext$ for the new task.

In all of our finetuning experiments we use the Adam optimizer \citep{kingma2014adam} and trained our models with a grid of hyperparameters: a learning rate of $2 \cdot 10^{-5}$ and $3 \cdot 10^{-5}$, for 3, 4, and 5 epochs with a linear learning rate decay, and two random seed for initialization.

To maximize performance on VCR, we also evaluate an ensemble of B2T2 models. Our ensemble is comprised of 5 identical B2T2 models, trained for 3 epochs with an initial learning rate of $2\cdot 10^{-5}$, but initialized with 5 different random seeds. The resulting class logits are then summed to obtain the ensemble scores.

%{\color{red} TODO: training details, here or in next section? Move this to next section?}

\section{Data}
\label{sec:data}

\emph{Visual Commonsense Reasoning} (VCR, \url{visualcommonsense.com}, \citealt{zellers2019}) is a corpus that contains a sample of stills from movies. Questions and answers revolve around conclusions or assumptions that require knowledge external to the images. The associated task is to not only select a correct answer but also provide reasoning in line with common sense.
Matching our problem formulation given before, a VCR sample is defined as a tuple $(I, O, Q, A, R)$.
Here, $I$ is the image, and $O$ is a sequence of objects identified in the image.
A question $Q = [q_0, \ldots, q_k]$ is given, where tokens are either textual words or deictic references to objects in $O$. 
Each question contains a set of four answers $A = \{A_1, A_2, A_3, A_4\}$, with exactly one correct answer $A^*$.
Each response follows the schema of the queries. Finally, there is a set of four rationales $R = \{R_1, R_2, R_3, R_4\}$, with exactly one rationale $R^*$ identified as correct in supporting $A^*$. 

Each of the objects in $O = [(b_1, l_1), \ldots, (b_{|O|}, l_{|O|})]$ is identified in the image $I$ by bounding boxes $b_i$.  The objects are also labeled with their classes with a text token $l_i$.

The $Q\rightarrow A$ task is to choose $A^*$ given $(I, O, Q, A)$.   The $QA\rightarrow R$ task is to choose $R^*$ given $(I, O, Q, A^*, R)$. Finally, the $Q \to AR$ task is a pipeline of the two, where a model must first correctly choose $A^*$ from $A$, then correctly choose $R^*$ given $A^*$.

% JL: this paragraph doesn't add much
% What makes VCR particularly interesting is that its creators focused on ``cognition-level questions,'' avoiding pitfalls brought on by some linguistic context (there are no unresolved referring expressions). Humans, with the benefit of additional contextual information from the movie scenes (rather than merely the still images) perform the tasks reasonably well (accuracy $Q\rightarrow A: 91\%; QA\rightarrow R: 93\%$). Models trained on VQA do not perform as well \citep{zellers2019}.

We adapt VCR to our problem formulation by converting each VCR example to four instances for the $Q \rightarrow A$ task, one per answer in $A$, and four instances for the $QA \rightarrow R$ task, one per rationale in $R$. We construct the text for the instances in the $Q \rightarrow A$ task as
\begin{align*}
  [&\mbox{[CLS]}, [b_0], q_0, \ldots, \mbox{[SEP]}, \\
   &a_0, \ldots, \mbox{[SEP]}, l_1, [b_1], \ldots, l_p, [b_p]]
\end{align*}
and in the $QA \rightarrow R$ task as
\begin{align*}
  [&\mbox{[CLS]}, [b_0], q_0, \ldots, \mbox{[SEP]}, a^*_0, \ldots, \\
  & r_0, \ldots, \mbox{[SEP]}, l_1, [b_1], \ldots, l_p, [b_p]].
\end{align*}
where $\mbox{[CLS]}, \mbox{[SEP]}$ are special tokens for BERT.

Here, $[b_0]$ is a bounding box referring to the entire input image. $q_0, \ldots$ are all question tokens, $a_0, \ldots$ answer tokens, $a^*_0, \ldots$ answer tokens for the correct answer, and $r_0, \ldots$ rationale tokens.
We append the first $p$ bounding boxes in $O$ with class labels to the end of the sequence (in our experiments, we use $p = 8$), and for objects referenced in $Q, A, R$, we prepend the class label token (i.e. $[b_i]$ becomes $l_i, [b_i]$).
We assign the binary label $l$ to every instance to represent whether the answer or rationale choice is the correct one.

% \subsection{GQA}

% \emph{GQA} is a dataset "for real-world visual reasoning and compositional question answering" (\url{visualreasoning.net}, \citealt{hudson2019gqa}). This set of 1.7M questions, compared to VQA, increases linguistic and semantic complexity in its questions and eliminates questions with strongly non-uniform distributed answers (which would be easy to answer by choosing the most likely answer).  While the VQA2.0 dataset may contain questions with more predictable answers, none of the datasets might not reflect the distribution of queries an actual user might make.  As \citet{hudson2019gqa} point out, a model trained on GQA generalizes well to the larger VQA dataset, and better so than in the reverse case.  

% Unlike VCR, GQA does not provide precomputed bounding boxes at evaluation time. We therefore preprocess every image in GQA with Mask-RCNN \citep{he2017mask} to extract bounding boxes, as well as an object label and an attribute label for every bounding box.
% {\color{red} TODO: say how Mask-RCNN was trained.}
% We extract 100 bounding boxes per example. We then construct the text for each instance as
% \[
%   [\mbox{[CLS]}, [b_0], t_1, \ldots, \mbox{[SEP]}, o_1, a_1, [b_1], \ldots, \mbox{[SEP]}].
% \]
% where $[b_0]$ is a bounding box for the entire image, $q_i$ are question tokens, $o_i$ are tokenized object labels, $a_i$ are tokenized attribute labels, and $[b_i]$ are references to each extracted bounding box.

\section{Experimental Results}

\subsection{VCR Task Performance}

Our final results on the VCR task are shown in Table \ref{tab:results}. Our Dual Encoder model worked surprisingly well compared to \citet{zellers2019}, surpassing the baseline without making use of bounding boxes.
We also evaluate a Text-Only baseline, which is similar to the Dual Encoder model but ignores the image.
The ensemble of B2T2 models, pretrained on Conceptual Captions, obtained absolute accuracy improvements of $8.9\%$, $9.8\%$ and $13.1\%$ compared to the published R2C baseline for the $Q \rightarrow A$, $QA \rightarrow R$, and $Q \rightarrow AR$ tasks respectively. At the time of this writing (May 22, 2019), both our single B2T2 and ensemble B2T2 models outperform all other systems in the VCR leaderboard.

% Suggest giving the relative reduction in error as well to be
% consistent with the abstract

\input{results-table}

\subsection{Ablations}
\label{sec:ablations}
\begin{table}[t]
    \small
    \centering
    \begin{tabular}{c|c}
         & $Q\to A$  \\
        \toprule
        Dual Encoder & 66.8 \\
        \midrule
        No bboxes & 67.5 \\
        Late fusion & 68.6 \\
        BERT-Base & 69.0 \\ 
        ResNet-50 & 70.4 \\
        No bbox class labels & 70.9 \\
        Fewer appended bboxes ($p=4$) & 71.0 \\
        No bbox position embeddings & 71.6 \\
        \midrule
        Full B2T2 & 71.9 \\
        \bottomrule
        % TODO: other ablations? position embeddings, number of bboxes
    \end{tabular}
    \caption{Ablations for B2T2 on VCR dev. The Dual Encoder and the full B2T2 models are the main models discussed in this work. All other models represent ablations from the full B2T2 model.}
    \label{tab:ablations}
\end{table}

To better understand the reason for our improvements, we performed a number of ablation studies on our results, summarized in Table \ref{tab:ablations}. We consider ablations in order of decreasing impact on the VCR dev set $Q \rightarrow A$ accuracy.

\begin{figure}
    \centering
    \includegraphics[width=\columnwidth]{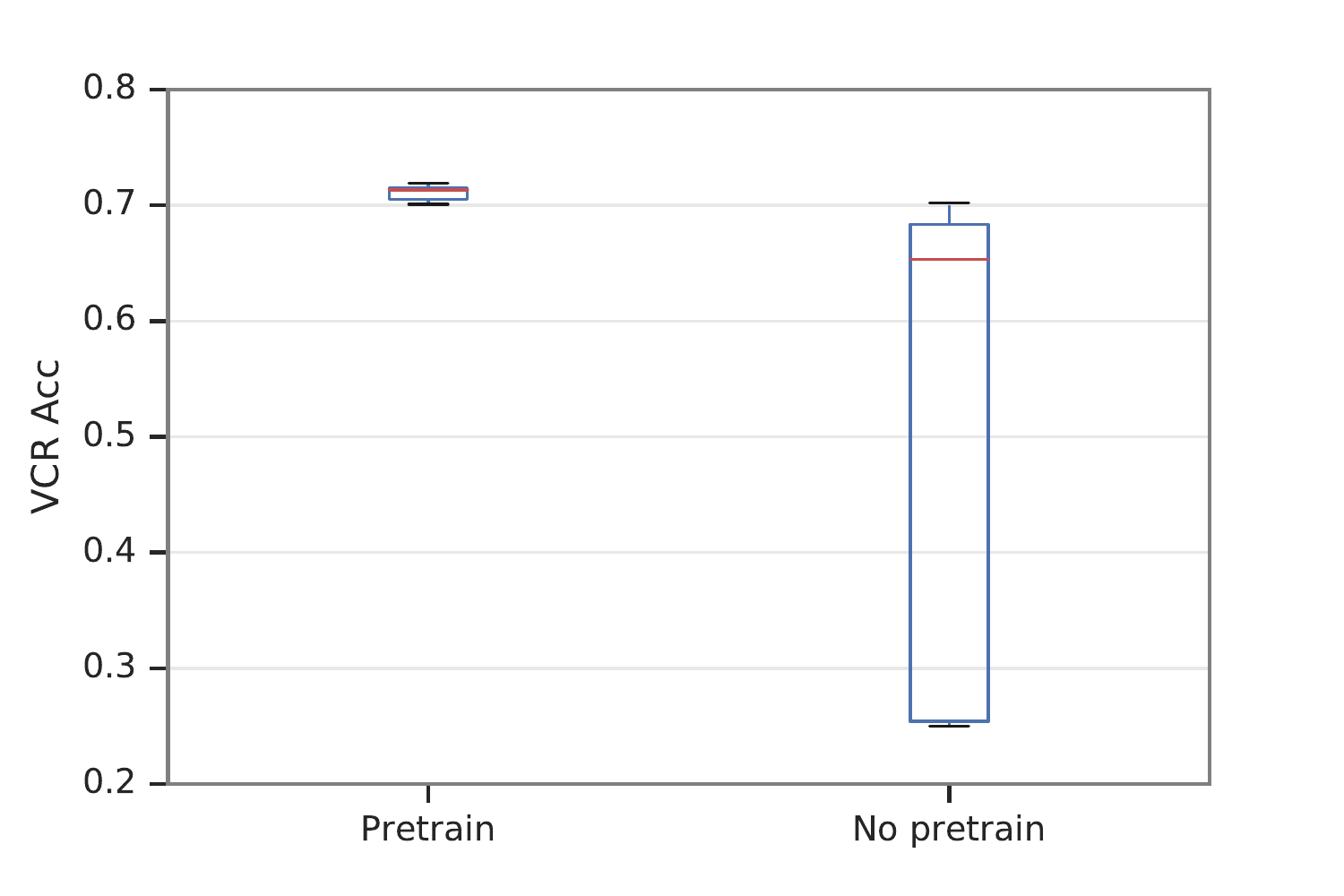}
    \caption{Boxplot of dev $Q \rightarrow A$ accuracy on VCR with and without pretraining. Pretraining on Conceptual Captions lowers variance when fine-tuning on VCR, from a grid search on multiple random seeds, learning rates, and VCR training epochs. }
    \label{fig:pretraining}
\end{figure}

{\bf Use of Bounding Boxes.} The bounding boxes considered by our model turns out to be the most important factor in improving the accuracy of our model. Without any bounding boxes we obtain $67.5\%$ accuracy, just above the accuracy of the dual encoder. With 4 instead of 8 appended bounding boxes we obtain $71\%$ accuracy. With 8 bounding boxes, but no textual labels from the bounding boxes in the text we obtain $70.9\%$ accuracy, showing that our model can make use of labels for detected objects. Example 1 in Table \ref{tab:examples} shows an example that our models can only get right if bounding box 5 is available.

{\bf Late Fusion vs.\ Early Fusion.} The second most important architectural choice in our model is to combine visual information at the level of context independent token embeddings, rather than at the highest levels of the neural representation. If in the the full B2T2 model we add visual embeddings in the last layer of BERT rather than in the first, we lose $3.3\%$ accuracy.

{\bf Effect of Textual Model Size.} The original VCR work by \citet{zellers2019} made use of BERT-base, while we use BERT-large to initialize our models. To test how much of our improvements are simply due to our model being larger, we retrained B2T2 models using BERT-base and found that we lose 2.9\% accuracy.

{\bf Effect of Visual Model Size.} How important is the choice of the visual model in the performance of B2T2? As further discussed in the error analysis section of this work, we suspect that B2T2 could be significantly improved by extending the visual features to represent more than just objects, but also activities, expressions and more. However it appears that even the size of the object detection model is important. If we swap out ResNet-152 for ResNet-50, accuracy decreases by $1.5\%$.

{\bf Pretraining.} We found that performance improvements from pretraining are quite small, around 0.4\% accuracy, but initializing from a pretrained model heavily reduces variance of results. We show this effect in Figure~\ref{fig:pretraining} over the grid of learning rates, random seeds, and training epochs described in Section \ref{sec:implementation}.

{\bf Position of Bounding Boxes} We additionally investigated the effect of removing position information from the model. The benefit of having bounding box positional embeddings is the smallest of the ones we considered. A model trained without positional embeddings only loses 0.3\% accuracy compared to the full model.

\subsection{Error Analysis}

\input{error-analysis}

\section{Related Work}

Modeling visual contexts can aid in learning useful sentence representations \citep{kiela2017learning} and even in training language models \citep{ororbia2019acl}. This paper takes these more general ideas to a downstream task that requires modeling of visual input.  Similar to B2T2, VideoBERT \citep{sun2019videobert} jointly processes video frames and text tokens with a Transformer architecture. However, VideoBERT cannot answer questions, nor does the model consider bounding boxes.

Our B2T2 model is similar to the Bottom-Up Top-Down attention model \citep{anderson2018bottom} in how bounding boxes generated at preprocessing time are attended to by the VQA model. ``Bottom-Up'' refers to the idea of attending from the text to the bounding boxes of objects detected in the image, while ``Top-Down'' refers to the idea of attending to regions constructed as a regular grid over the image. The Bottom-Up Top-Down model however reduces the text to a fixed length vector representation before attending to image regions, while B2T2 instead treats image regions as special visual tokens mixed in the text. In this sense, Bottom-Up Top-Down model is a late fusion model, while B2T2 is early fusion.

The Neuro-Symbolic Concept Learner \citep{mao2019neuro} also uses bounding boxes to learn visually grounded concepts through language. The Neuro-Symbolic Concept Learner however relies on a semantic parser to intepret language, while B2T2 
uses a Transformer to construct a joint representation of textual tokens and visual tokens.

Another recently proposed model for VQA is MAC \citep{hudson2018compositional}. As presented, MAC does not make use of bounding boxes, which makes it a Top-Down model in the nomenclature of \citet{anderson2018bottom}. MAC also reduces the textual information to a vector of fixed length. However MAC makes use of a new neural architecture designed to perform an explicit multi-step reasoning process and is reported to perform better than \citet{anderson2018bottom} on the GQA dataset \citep{hudson2019gqa}.

After the submission of this paper, several new works were published with excellent results on VCR, in some cases exceeding the performance of our system. In particular we mention ViLBERT \citep{lu2019vilbert}, VL-BERT \citep{su2019vl}, Unicoder-VL \citep{li2019unicoder}, and VisualBERT \citep{li2019visualbert}.

VCR is only one of several recent datasets pertaining to the visual question answering task. VQA \citep{VQA, balanced_binary_vqa, balanced_vqa_v2} contains photos and abstract scenes with questions and several ground-truth answers for each, but the questions are less complex than VCR's. CLEVR \citep{johnson2017clevr} is a visual QA task with compositional language, but the scenes and language are synthetic. GQA \citep{hudson2019gqa} uses real scenes from Visual Genome, but the language is artificially generated.
%The VQA v2 dataset \citep{balanced_vqa_v2} contains about 265,000 images, an average of $5.4$ questions per image and $10$ ground truth answers to each question.
Because VCR has more complex natural language than other datasets, we consider it the best evaluation of a model like B2T2, which has a powerful language understanding component. 

\section{Conclusion}

In this work we contrast different ways of combining text and images when powerful text and vision models are available. We picked BERT-Large \cite{devlin2018bert} as our text model, ResNet-152 \cite{he2016identity} as our vision model, and the VCR dataset \cite{zellers2019} as our main benchmark.

The early-fusion B2T2 model, which encodes sentences along with links to bounding boxes around identified objects in the images, produces the best available results in the visual question answering tasks.  A control model, implementing late fusion (but the same otherwise), performs substantively worse.  Thus, grounding words in the visual context should be done early rather than late.

We also demonstrate competitive results with a Dual Encoder model, matching state-of-the-art on the VCR dataset even when textual references to image bounding boxes are ignored.  We then showed that our Dual Encoder model can be substantially improved by deeply incorporating in the textual embeddings visual features extracted from the entire image and from bounding boxes.  We finally show that pretraining our deep model on Conceptual Captions with a Mask-LM loss yields a small additional improvement as well as much more stable fine-tuning results.

\bibliography{visual-qa}
\bibliographystyle{acl_natbib}

\end{document}

%% file: results-table.tex
% THIS IS THE RESULTS TABLE COPIED FROM ZELLERS ET AL!

\newcommand{\nodata}[1]{\multicolumn{#1}{|c|}{\cellcolor{gray!10}}}
\newcommand{\nodataright}[1]{\multicolumn{#1}{|c}{\cellcolor{gray!10}}}
\newcommand{\nd}{\cellcolor{gray!10}}

\newcommand{\tinyrule}{\cline{2-3} \\[-1.0em]}

\newcommand{\resultswidth}{1.35cm}
\begin{table}[t!]
\vspace{-3mm}
\centering
\resizebox{\columnwidth}{!}{%
\begin{small}
\setlength{\tabcolsep}{4pt}
\begin{tabular}{@{} c@{\hspace{0.4em}} l @{\hspace{0.7em}}|
c@{\hspace{0.7em}}c@{\hspace{0.5em}} |c@{\hspace{0.7em}}c  |c@{\hspace{0.7em}}c@{}}
&\multicolumn{1}{c}{} & \multicolumn{2}{c}{$Q \rightarrow A$} & \multicolumn{2}{c}{$QA \rightarrow R$} & \multicolumn{2}{c}{$Q \rightarrow AR$}\\ 
\multicolumn{2}{c|}{Model} & Val & Test & Val & Test & Val & Test \\ 
\toprule
& Chance & 25.0 & 25.0 & 25.0 & 25.0 & \phantom{0}6.2 &  \phantom{0}6.2 \\ %\spacedhline
\midrule
& Text-Only BERT (Zellers et al.) & 53.8 & 53.9 & 64.1 & 64.5 & 34.8 & 35.0 \\ 
%\multirow{4}{*}{\rotatebox[origin=c]{90}{Text Only}}& BERT & 53.8 & 53.9 & 64.1 & 64.5 & 34.8 & 35.0 \\ 
%& BERT (response only) & 27.6 & 27.7 & 26.3 & 26.2 & \phantom{0}7.6 & \phantom{0}7.3 \\ 
%& ESIM+ELMo & 45.8 & 45.9 & 55.0 & 55.1 & 25.3 & 25.6 \\ 
%& LSTM+ELMo & 28.1 & 28.3 &  28.7 & 28.5 & \phantom{0}8.3 & \phantom{0}8.4 \\ \midrule

%%\multirow{4}{*}{\rotatebox[origin=c]{90}{VQA}} 

% SOME UNPUB RESULTS?
% & BottomUp+BERT & \multicolumn{2}{c}{??} & 63.0 & 62.9 & \multicolumn{2}{c}{??} \\ 

%& RevisitedVQA \cite{jabri2016revisiting} & 39.4 & 40.5 & 34.0 & 33.7 & 13.5 & 13.8 \\ 
%& BottomUpTopDown\cite{Anderson2017updown} & 42.8 & 44.1 & 25.1 & 25.1 & 10.7 & 11.0 \\  
%& MLB \cite{Kim2017} & 45.5 & 46.2 & 36.1 & 36.8 & 17.0 & 17.2 \\ 
%& MUTAN \cite{Ben-younes_2017_ICCV} & 44.4 & 45.5 & 32.0 & 32.2 & 14.6 & 14.6 \\ \midrule
%% \\ \midrule
& R2C (Zellers et al.) & 63.8 & 65.1 & 67.2 & 67.3 & 43.1 & 44.0 \\ \midrule
%& FAIR (unpub.) & - & 65.7 & - & 70.1 & - & 46.3 \\
%& UTS SGRE (unpub.) & - & 67.5 & - & 69.5 & - & 46.9 \\
% & MUGRN (subm. NeurIPS) & - & 68.2 & - & 69.4 & - & 47.5 \\
% & SNU MRCNet (unpub.) & - & 68.4 & - & 70.5 & - & 48.4 \\
% & UTS CCD (unpub.) & - & 68.5 & - & 70.5 & - & 48.4 \\
& HCL HGP (unpub.) & - & 70.1 & - & 70.8 & - & 49.8 \\
& TNet (unpub.) & - & 70.9 & - & 70.6 & - & 50.4 \\
& B-VCR (unpub.) & - & 70.5 & - & 71.5 & - & 50.8 \\
& TNet 5-Ensemble (unpub.) & - & 72.7 & - & 72.6 & - & 53.0 \\
\midrule
& Text-Only BERT (ours) & 59.5 & - & 65.6 & - & 39.3 & - \\ 
& Dual Encoder (ours) & 66.8 & - & 67.7 & - & 45.3 & - \\
& B2T2 (ours) & 71.9 & 72.6 & 76.0 & 75.7 & 54.9 & 55.0 \\
& B2T2 5-Ensemble (ours) & {\bf 73.2} & \bf 74.0 & \bf 77.1 & \bf 77.1 & \bf 56.6 & \bf 57.1 \\
\midrule
% & One turker & & 86.7 & & 87.6 & & 76.0 \\ 
% & Three turkers & & 90.4 & & 91.2 & & 82.9 \\ 
& Human & & 91.0 & & 93.0 & & 85.0 \\ 
\bottomrule
\end{tabular}
\end{small}}
\vspace*{-1mm}\caption{Experimental results on VCR, incorporating those reported by \citet{zellers2019}.  The proposed B2T2 model and the B2T2 ensemble outperform published and unpublished/undocumented results found on the VCR leaderboard at \url{visualcommonsense.com/leaderboard} as of May 22, 2019. %: SGRE may be based on \citet{wu2018}.
} \vspace*{-1mm}
\label{tab:results}
\end{table}

% Table with GQA results.

% \begin{table}[t!]
% \vspace{-3mm}
% \centering
% \begin{small}
% \setlength{\tabcolsep}{4pt}
% \begin{tabular}{c|cc}
%  & \multicolumn{2}{c}{Accuracy} \\
%  & Dev & Test \\ \midrule
% Bottom-Up &  &  49.74 \\
% MAC &  &  54.06 \\ \midrule
% {\bf Deep Bottom-Up} & 54.87 &  \\ \midrule
% LXR955 &  & 60.34 \\  \midrule
% Human &  & 89.3 \\ 
% \bottomrule
% \end{tabular}
% \end{small}
% \vspace*{-1mm}\caption{Experimental results on GQA. Our model compares favorably to the baselines Bottom-Up and MAC described in \citep{hudson2019gqa}. LXR955 is an unpublished result that as of this writing is at the top of the GQA leaderboard. {\color{red} TODO: this is still training.}}\vspace*{-1mm}
% \label{tab:gqa-results}
% \end{table}

%% file: error-analysis.tex
\begin{table*}
\begin{footnotesize}
{\setlength{\tabcolsep}{2pt}
\begin{tabular}{cc}

% Example 1
\parbox[c]{9cm}{\includegraphics[width=9cm]{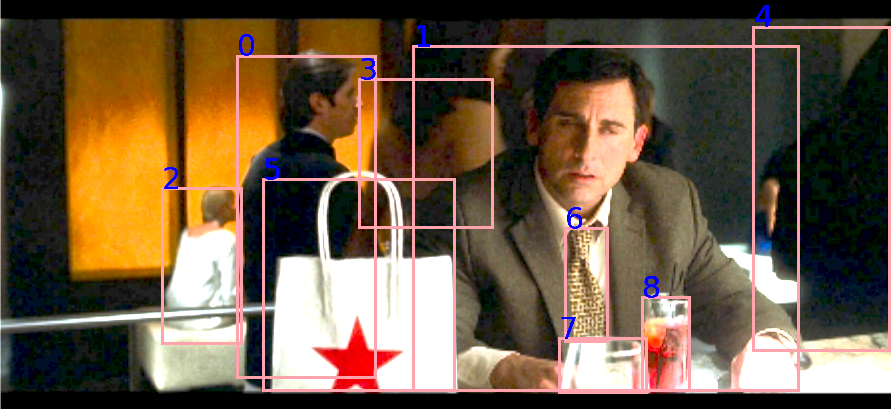}} \vspace{2mm}
&
\begin{footnotesize}
{\setlength{\tabcolsep}{2pt}
\begin{tabular}{cp{6.5cm}} \\
\multicolumn{2}{l}{Example 1 \vspace{1mm}}\\
Q: & What did [1] do before coming to this location? \vspace{1mm} \\
A$_1$: & He took horse riding lessons. (text-only) \\
\textbf{A$_2$:} & \textbf{He was just shopping. (B2T2)} \\
A$_3$: & He found a skeleton. \\
A$_4$: & He came to buy medicine. (dual encoder)
\end{tabular}}
\end{footnotesize} \\

% Example 2
\parbox[c]{9cm}{\includegraphics[width=9cm]{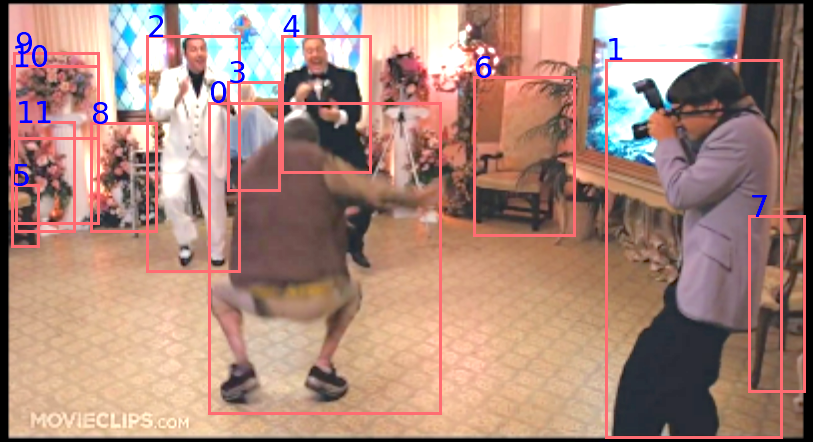}} \vspace{2mm}
&
\begin{footnotesize}
{\setlength{\tabcolsep}{2pt}
\begin{tabular}{cp{6.5cm}} \\
\multicolumn{2}{l}{Example 2 \vspace{1mm}}\\
Q: & How are [2, 4] related? \vspace{1mm} \\
A$_1$: & [2, 4] are partners on the same mission. \\
\textbf{A$_2$:} & \textbf{[2, 4] are a recently married gay couple. (B2T2)} \\
A$_3$: & They are likely acquaintances. \\
A$_4$: & They are siblings. (text-only, dual encoder)
\end{tabular}}
\end{footnotesize} \\

% Example 3
\parbox[c]{9cm}{\includegraphics[width=9cm]{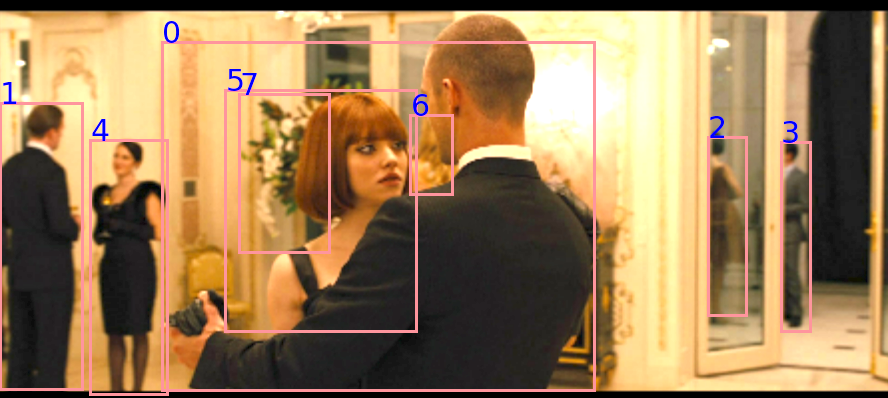}}  \vspace{2mm}
&
  \begin{footnotesize}
{\setlength{\tabcolsep}{2pt}
\begin{tabular}{cp{6.5cm}} \\
\multicolumn{2}{l}{Example 3 \vspace{1mm}}\\
Q: & What are [0] and the woman doing? \vspace{1mm} \\
A$_1$: & Their husbands are doing something dumb. \\
A$_2$: & They are observing the results of an experiment.  (text-only, dual encoder) \\
\textbf{A$_3$:} & \textbf{They are dancing. (B2T2)} \\
A$_4$: & They are acting as nurses for the rescued people.
\end{tabular}}
  \end{footnotesize} \\

% Example 4
  \parbox[c]{9cm}{\includegraphics[width=9cm]{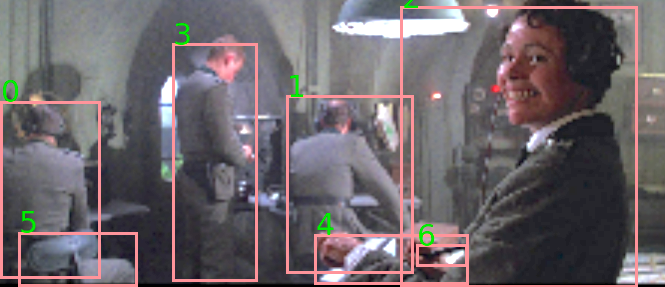}} \vspace{2mm}
  &
  \begin{footnotesize}
{\setlength{\tabcolsep}{2pt}
\begin{tabular}{cp{6.5cm}} \\
\multicolumn{2}{l}{Example 4 \vspace{1mm}}\\
Q: & How is [2] feeling? \vspace{1mm} \\
A$_1$: & [2] is feeling shocked. (B2T2, dual encoder) \\
A$_2$: & [0] is feeling anxious. \\
A$_3$: & [2] is not feeling well. \\
\textbf{A$_4$:} & \textbf{[2] is feeling joy and amusement. (text-only)}
\end{tabular}}
  \end{footnotesize}
\\

% Example 5
\parbox[c]{9cm}{\includegraphics[width=9cm]{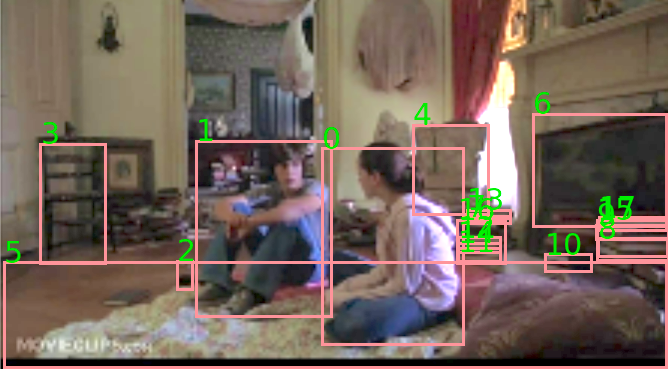}}
&
\begin{footnotesize}
{\setlength{\tabcolsep}{2pt}
\begin{tabular}{cp{6.5cm}} \\
\multicolumn{2}{l}{Example 5 \vspace{1mm}}\\
Q: & Why is [1] on the floor talking to [0]? \vspace{1mm} \\
A$_1$: & The man on the floor was assaulting [1]. \\
A$_2$: & He is asking her to help him stand up. (B2T2, dual encoder) \\
A$_3$: & [1] just dropped all his books on the floor. \\
\textbf{A$_4$:} &\textbf{[1] looks like he is telling [0] a secret. (text-only)}
\end{tabular}}
  \end{footnotesize}  \\
  
\end{tabular}}
\end{footnotesize}
\caption{Examples of the $Q \rightarrow A$ task from the VCR dev set. The correct answer for every example is marked in bold. The answers picked by the text-only model, by the dual encoder and by B2T2 are indicated in parenthesis.}
\label{tab:examples}
\end{table*}

We picked some examples, shown in Table \ref{tab:examples}, to illustrate the kinds of correct and incorrect choices that B2T2 is making, compared to our dual encoder and to a text only model.

In Example 1 we show an example of how our model picks the right answer only when it is able to make use of all provided bounding boxes. Bounding box 5 in particular contains the clue that allows the observer to know that the man in the picture might have just gone shopping.

In Examples 2 and 3, no specific bounding box appears to contain critical clues for answering the question, but B2T2 outperforms models without access to the image or without access to bounding boxes. It is possible that B2T2 might be gaining deeper understanding of a scene by combining information from important regions of the image.

In Examples 4 and 5, we see failure cases of both the dual encoder and B2T2 compared to the text only-model. Both these examples appear to point to a limitation in the amount of information that the we are able to extract from the image. Indeed our vision model is trained on ImageNet, and so it might be very good at recognizing objects, but might be unable to recognize human expressions and activities. Our models could have correctly answered the question in Example 4 if they were able to recognize smiles. Similarly our models could have ruled out the incorrect answer they picked for the question in Example 5 if they were able to see that both people in the picture are sitting down and are not moving.